\documentclass[journal]{IEEEtranTIE}
\usepackage{graphicx}
\usepackage{cite}
\usepackage{picinpar}
\usepackage{amsmath}
\usepackage{url}
\usepackage{flushend}
\usepackage[latin1]{inputenc}
\usepackage{colortbl}
\usepackage{soul}
\usepackage{multirow}
\usepackage{pifont}
\usepackage{color}
\usepackage{alltt}
\usepackage[hidelinks]{hyperref}
\usepackage{enumerate}
\usepackage{siunitx}
\usepackage{breakurl}
\usepackage{epstopdf}
\usepackage{pbox}
\usepackage{multicol}
\usepackage{multirow} 
\newcommand{\jun}[1]{{\color{black}#1}} 
\newcommand{\tabincell}[2]{\begin{tabular}{@{}#1@{}}#2\end{tabular}}
\usepackage{booktabs}

\begin{document}
\title{Fast and Accurate Normal Estimation for Point Cloud via Patch Stitching}

\author{
	\vskip 1em
	
	Jun Zhou$^*$,
	Wei Jin,
	Mingjie Wang,
	Xiuping Liu,
	Zhiyang Li,
	Zhaobin Liu
	\thanks{
		
		
	
	This work is supported by the National Natural Science Foundation of China (No. 62002040, No.61976040).
	
	J. Zhou, Z. Li, Z. Liu are with the School of information science and technology, Dalian Maritime University, China
	
	W. Jin is with Dalian Neusoft University of Information,China.
	
	M. Wang is with Memorial University of Newfoundland, Canada.

	X. Liu is with with Dalian University of Liaoning Province, China.
	
	E-mail: jun90@dlmu.edu.cn
	}
}

\maketitle
	
\begin{abstract}
This paper presents an effective normal estimation method adopting multi-patch stitching for an unstructured point cloud. The majority of learning-based approaches encode a local patch around each point of a whole model and estimate the normals in a point-by-point manner. In contrast, we suggest a more efficient pipeline, in which we introduce a patch-level normal estimation architecture to process a series of overlapping patches. Additionally, a multi-normal selection method based on weights, dubbed as multi-patch stitching, integrates the normals from the overlapping patches.
To reduce the adverse effects of sharp corners or noise in a patch, we introduce an adaptive local feature aggregation layer to focus on an anisotropic neighborhood. We then utilize a multi-branch planar experts module to break the mutual influence between underlying piecewise surfaces in a patch.
At the stitching stage, we use the learned weights of multi-branch planar experts and distance weights between points to select the best normal from the overlapping parts. Furthermore, we put forward constructing a sparse matrix representation to reduce large-scale retrieval overheads for the loop iterations dramatically. Extensive experiments demonstrate that our method achieves SOTA results with the advantage of lower computational costs and higher robustness to noise over most of the existing approaches.
\end{abstract}


\markboth{}%
{}

\definecolor{limegreen}{rgb}{0.2, 0.8, 0.2}
\definecolor{forestgreen}{rgb}{0.13, 0.55, 0.13}
\definecolor{greenhtml}{rgb}{0.0, 0.5, 0.0}

\section{Introduction}
\IEEEPARstart{D}{ue}  to the advances in 3D acquisition technologies, point-based representations can be captured in various vision applications. Generally, the scanned large-scale point cloud is noisy, incomplete, and contains sampling irregularity while lacking the essential local geometry properties such as point normals. An accurate and fast normal estimation algorithm can significantly improve the downstream tasks such as 3D surface reconstruction \cite{berger2014state}, point cloud consolidation \cite{wu2015deep,wang2013consolidation}.

\begin{figure}[h]
 \centering 
 \includegraphics[width= .95\columnwidth]{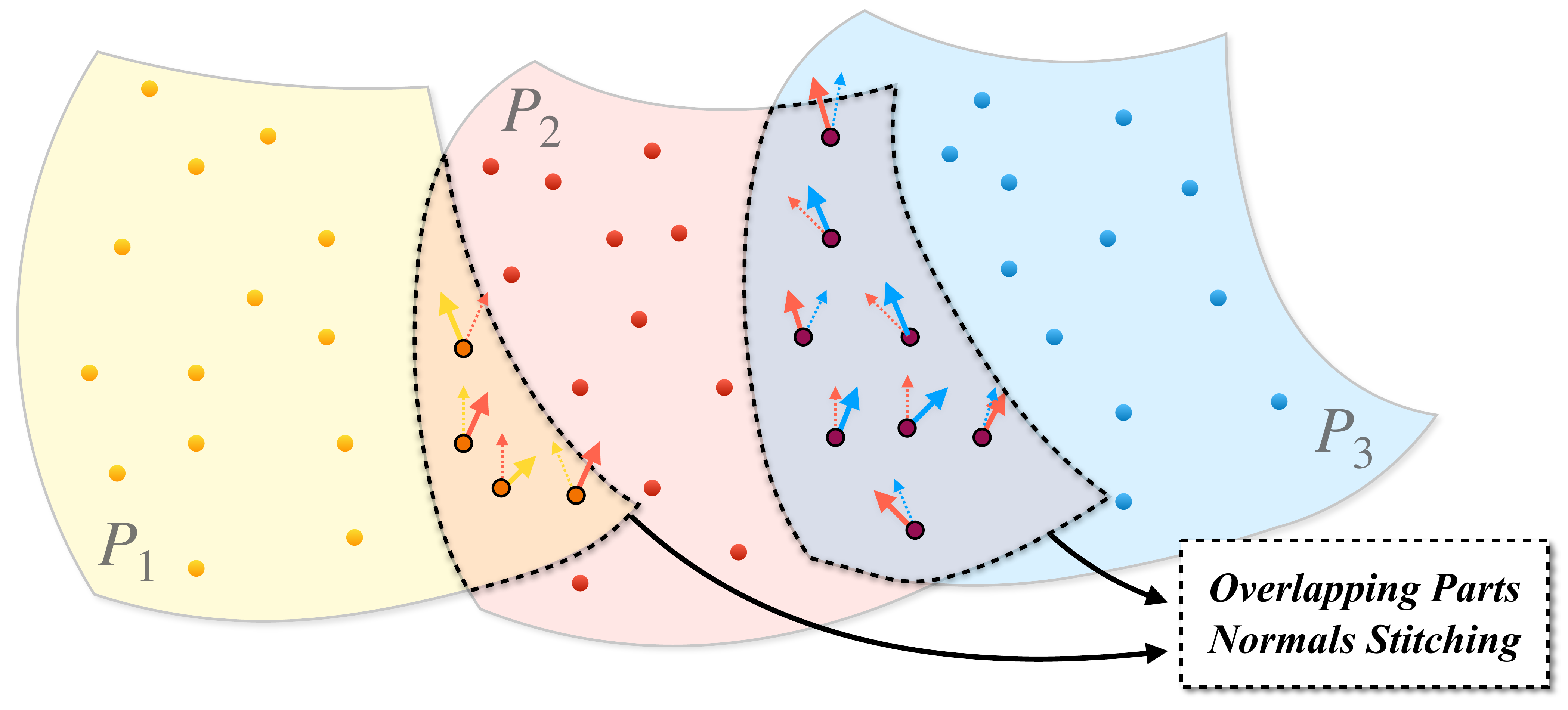}
 \caption{The illustration of multi-patch (normals) stitching.}

 \label{fig:5}
\end{figure}

Compared with traditional methods, the learning-based methods \cite{guerrero2018pcpnet,ben2019nesti,ben2020deepfit} can well handle the problem of parameter tuning. Commonly, given a point, neighboring points should be extracted via K nearest neighbor (kNN) search and used to infer the normal of the center point via the deep learning architecture. In this process, the selection of neighborhood size is greatly affected by noise. The multi-scale strategy may achieve higher accuracy but suffers from high computation time. Besides,  assuming that a model contains 100k points and 256 neighbors are sampled by each point, a large repetition rate would occur (Each point would be used about 256 times evenly), leading to a computational redundancy. Although the IterNet \cite{lenssen2020deep} can quickly implement normal estimation with fewer parameters, the whole model needs to be input simultaneously, making the algorithm inflexible.

\begin{figure*}[h]
 \centering 
 \includegraphics[width= 0.95\textwidth]{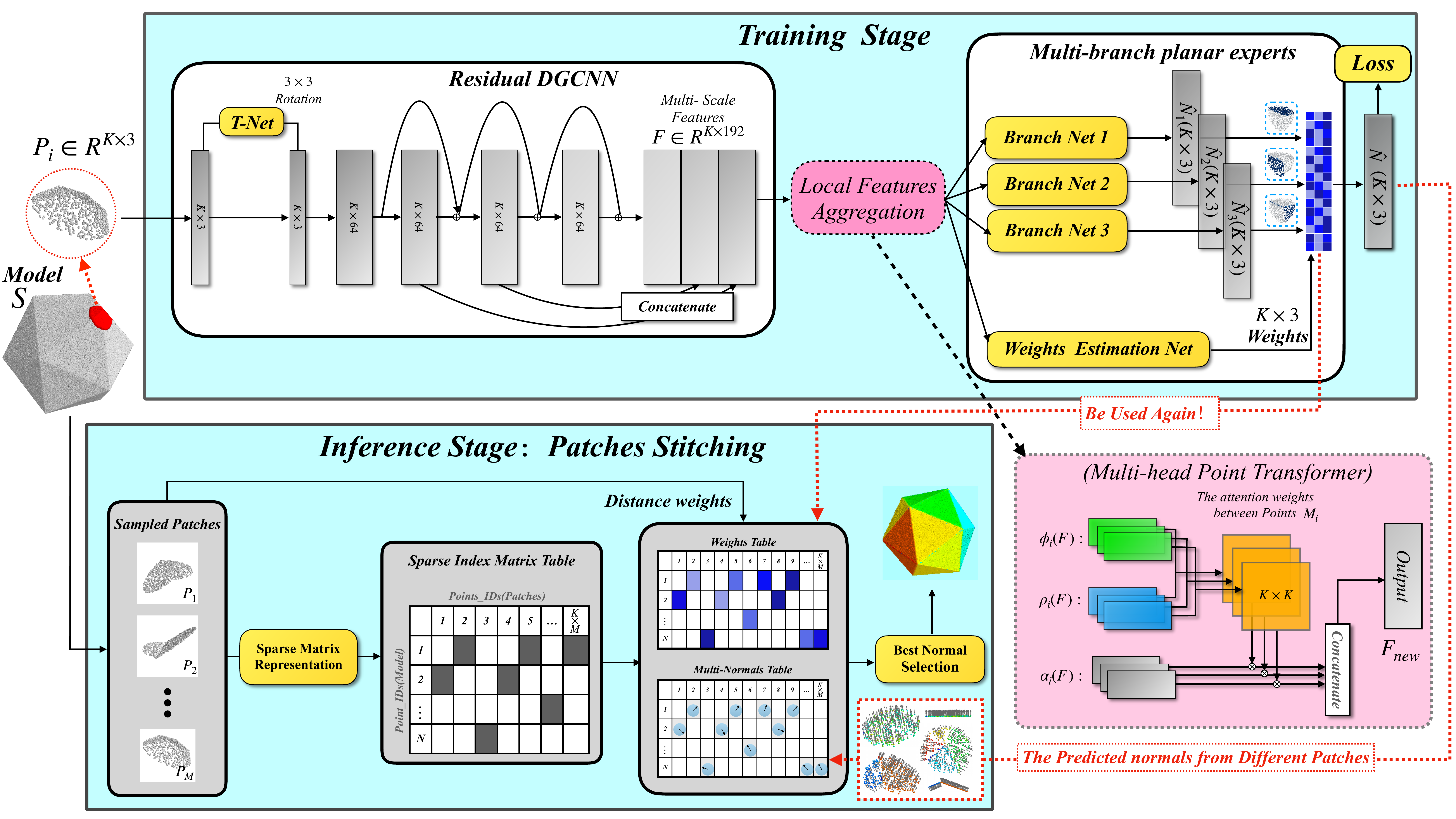}
 \caption{The pipeline of our normal estimation method. Given a point cloud $S$, we employ the trained network to estimate patch-level normals from the sampled patches, and the weights of multi-branch planar experts can be obtained simultaneously. In the stage of patch stitching, a sparse index representation is constructed to map the points \textit{ID}s of the overlapping patches to the \textit{ID}s of the point cloud. The learned experts weights and the patch distance weights are used to select the best normal from the overlapping parts with the same \textit{ID}s.}
 \label{fig:1}
\end{figure*}

Motivated by these challenges, we adopt a patch-by-patch normal estimation method to effectively reduce the repetition rate of each point on the model without sacrificing the size of the sampling range, and only a few patches are used for whole model estimation (see Fig.~\ref{fig:5}). A sparse index matrix is constructed to map the point $ID$s of the patches to the entire point cloud, significantly reducing the time-cost of the multi-normal stitching process. Besides, large-scale sampling patches with isotropy may contain underlying piecewise surfaces and noise. Thus, the useless information within neighbors and mutual interference between underlying surfaces will lead to entanglement between points at the patch-level normal estimation stage. To solve the entanglement problem, we firstly employ an adaptive local feature aggregation layer to focus on a meaningful neighborhood for each point at the feature extraction stage. Then a multi-branch planar experts module is introduced to break down the mutual interference between the underlying surfaces at the normal estimation stage. The contributions of our work are summarized as follows: 
\begin{itemize}
\item A flexible and fast multi-patch stitching framework is proposed, and the network's time complexity can be ignored relative to the scale of point cloud.
\item A sparse index matrix is introduced to speed up the process that maps the point \textit{ID}s of patches to the point cloud. The mapping table can be used at the stage of weight-based normals selection.
\item  A multi-branch planar experts module and an adaptive local feature aggregation layer are introduced to greatly improved the robustness of our patch-level network. 
\end{itemize}

 \begin{figure*}[ht]
 \centering 
 \includegraphics[width= 0.95\textwidth]{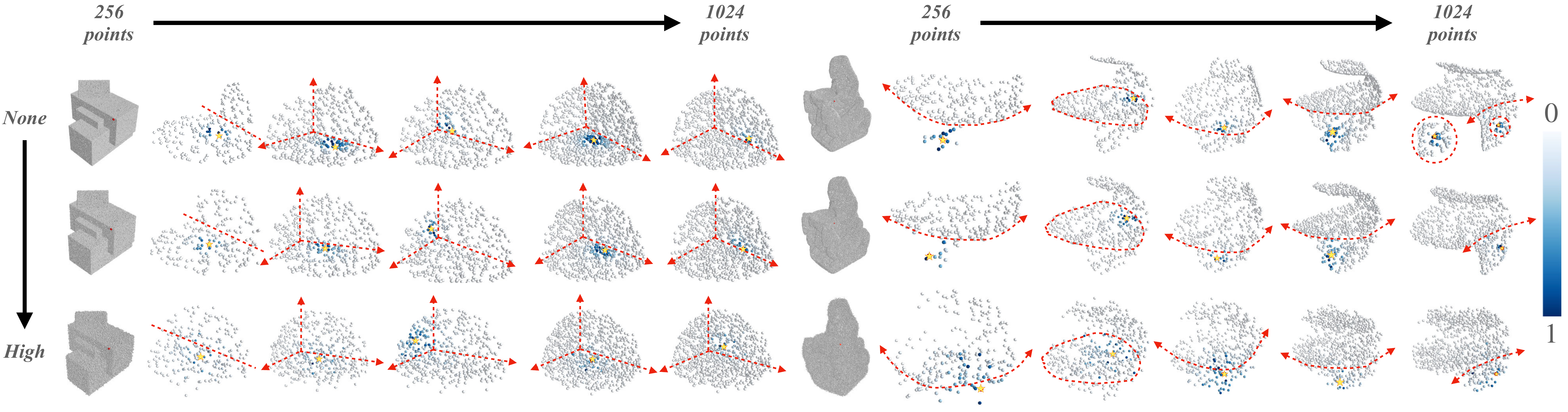}
 \caption{Visualization of attention weights sampled from two kinds of models. From top to bottom: patches without noise, 0.012 Gaussian noise, and 0.006 Gaussian noise. For each model, from left to right: patches with different sampling sizes. Patches are colored according to weight magnitude mapped to a heatmap ranging from 0 to 1. }
 \label{fig:2}
\end{figure*}

\section{Related Work}
The task of normal estimation for point cloud is a long-standing fundamental problem in geometry processing. In this section, we will review traditional and popular deep-based methods for normal estimation. 

\subsection{Traditional normal estimation}
A classic regression method for normal estimation uses principal component analysis (PCA)~\cite{hoppe1992surface}. However, the performance of this approach usually depends on the patch sizes and the noise scales. Thus, some more effective sampling methods~\cite{mitra2003estimating,pauly2003shape} are proposed. Several variants are also proposed for plane fitting,  such as local spherical surfaces fitting~\cite{guennebaud2007algebraic} and Jets~\cite{cazals2005estimating}.  \jun{However, the effect of sampling size is still existed in these methods, in which the isotropic neighborhoods (which always contain different underlying piecewise surfaces) will affect the final fitting process.}  
For better discarding the neighbors belonging to different surfaces on the patch, more robust statistics approaches are employed to select an adaptive region~\cite{li2010robust,mederos2003robust,yoon2007surface,wang2013adaptive}. While many of these methods hold theoretical guarantees on approximation and robustness, all the above methods require a careful setting of parameters for different shape types. So the data-driven methods are essentially necessary technologies to improve the robustness of the point cloud normal estimation.

\subsection{Learning-based normal estimation}
Several deep learning approaches have been proposed to estimate normal vectors from unstructured point clouds in recent years. Commonly, the existing deep learning strategy is always inspired by the traditional methods: a local neighborhood should be selected first, and then a regressor is proposed to evaluate the normal of a center point. As a pioneer, Boulch and Marlet~\cite{boulch2016deep} first apply 2D CNN to regress the point normal. Then, inspired by PointNet~\cite{guerrero2018pcpnet}, PCPNet~\cite{guerrero2018pcpnet} directly uses local neighboring points to regress the point cloud normal. Besides, based on 3D modified fisher vectors (3DmFV)~\cite{ben20183dmfv} presentations, Nesti-Net~\cite{ben2019nesti} is proposed to use a mixture-of-experts architecture, which relies on a data-driven approach for selecting the optimal scale around each point and encourages sub-network specialization. The multi-scale methods can effectively improve the results, but it caused a great time consumption. Recently, Zhou et al.~\cite{zhou2020normal} introduce an extra feature constraint mechanism that can effectively distinguish the piecewise surfaces near the sharp edges of a patch. Then Deepfit~\cite{ben2020deepfit} incorporates a neural network to learn pointwise weights for weighted least squares surface fitting and make a significant performance. However, these methods also follow the pipeline of PCPNet, which evaluates normal in a point-by-point way. Lenssen et al. ~\cite{lenssen2020deep} employ an iteration normal estimation method to reduce time consumption. This method has to input the entire model, thus reducing the flexibility of the network. Considering the time-cost and the flexibility of the network, we propose a fast patch stitching method for normal estimation. Our method can achieve normal estimation more efficiently by a stitching way (see Fig.~\ref{fig:5}) than most existing deep learning methods regardless of network's time complexity. 

\section{Overview}
This paper presents a novel pipeline that uses a multi-patch stitching strategy to accelerate the process of normal estimation (see Fig.~\ref{fig:1}). It consists of the following two main parts:  patch-level normal estimation and multi-patch normal stitching. Firstly, a residual DGCNN (Sec.~\ref{sec:Basic_Net}) is proposed as our patch-level network, which incorporates an \jun{adaptive local feature aggregation layer} (Sec.~\ref{sec:Features_Aggregation}) and a \jun{multi-branch planar experts module} (Sec.~\ref{sec:Planar_Experts}). Then, a \jun{sparse index matrix} is built to efficiently map the point \textit{ID}s of patches to the whole point cloud in the multiple patches stitching (Sec.~\ref{sec:Patch_Stitching}). Finally, evaluating weight is used to select the best normal from the overlapping patches.

\section{Patch-level normal estimation} 
\subsection{Pre-processing}
The full pipeline is illustrated in Fig.~\ref{fig:1}. Given a 3D point cloud $S \in R^{N \times 3}$ and a query point $p_{i} \in S$, a local patch $P_{i} \in R^{K \times 3}$ can be extracted via \jun{kNN}. In the training stage, we randomly sample patches from the models to ensure sufficient training samples. In the inference stage, the $M$ patches that cover the entire model $S$ are extracted by using the farthest point sampling (FPS) algorithm and $M \ll N$.

\subsection{The Basic network: residual DGCNN}\label{sec:Basic_Net}
Given a sampled patch $P_{i}  \in R^{k\times 3}$ as input, our estimator can predict each point normal of the patch. Therefore, we design a basic network inspired by DGCNN~\cite{wang2019dynamic}, which can extract the pointwise feature of a patch and a simple regressor can be followed to estimate the pointwise normals. Firstly, \jun{a quaternion spatial transformer (QST)} is used to transform the input patch to a canonical pose $\hat{P} = P \cdot T_{\Theta_{t}}(P)$, and $T_{\Theta_{t}}(P)$ is a learned rigid rotation transformation. Then, four stacked graph convolution layers are used to extract pointwise features of the rotated patch $\hat{P}$.

Specifically, in the first layer, we construct a directed graph $\mathcal{G}^{0}=(\mathcal{V}^{0}, \mathcal{E}^{0})$ where $\mathcal{V}^{0}$ and $\mathcal{E}^{0}$ are unordered point set $\{\hat{p}_{i}\}_{i=1}^{K}$ and \jun{kNN} graph of the patch $\hat{P}$, respectively. We define edge features of the $0$-th layer as $\textbf{e}_{ij}^{0} = h_{\Theta_{0}}(\hat{p}_{i},\hat{p}_{j})$, where $h_{\Theta_{0}}$ is 
\begin{equation}
h_{{\Theta}_{0}} = ReLU(\theta_{0}\cdot (\hat{p}_{j}-\hat{p}_{i})+\bar{\theta}_{0}\cdot \hat{p}_{i}).  
\end{equation}
Here, $(\hat{p}_{i},\hat{p}_{j}) \in  \mathcal{E}^{0}$, $\Theta_{0} = (\theta_{0},\bar{\theta}_{0})$, and the output feature $F^{0}\in R^{K\times 64}$ at the vertex $i$ is thus given by
\begin{equation}
    f_{i}^{0} =  \max_{j:(i,j)\in  \mathcal{E}^{0}} h_{\Theta_{0}}(\hat{p}_{i},\hat{p}_{j}).
\end{equation}
From the second to forth layer of the basic network, a residual map is learned to take feature  $F^{l}\in R^{K\times 64}$ as input and output residual feature representation $F^{l+1}$ for the next layer. The learned feature of vertex $i$ can be  defined as
\begin{equation}
    f_{i}^{l+1} =  \max_{j:(i,j)\in \mathcal{E}^{l+1}} h_{\Theta_{l+1}}(f^{l}_{i},f^{l}_{j}) +f_{i}^{l},
\end{equation}
where $l=0,1,2,3$, and the $\mathcal{E}^{l+1}$ is updated via a dynamic graph strategy~\cite{wang2019dynamic}. Finally, we concatenated the multi-scale features as global feature $F= F^{1}\bigoplus F^{2} \bigoplus F^{3} \in R^{K \times 192}$, and $\{\Theta_{t},\Theta_{0},\Theta_{1},\Theta_{2},\Theta_{3}\}$ are the training parameters of the basic network.

\begin{figure*}[th]
 \centering 
 \includegraphics[width=0.95\textwidth]{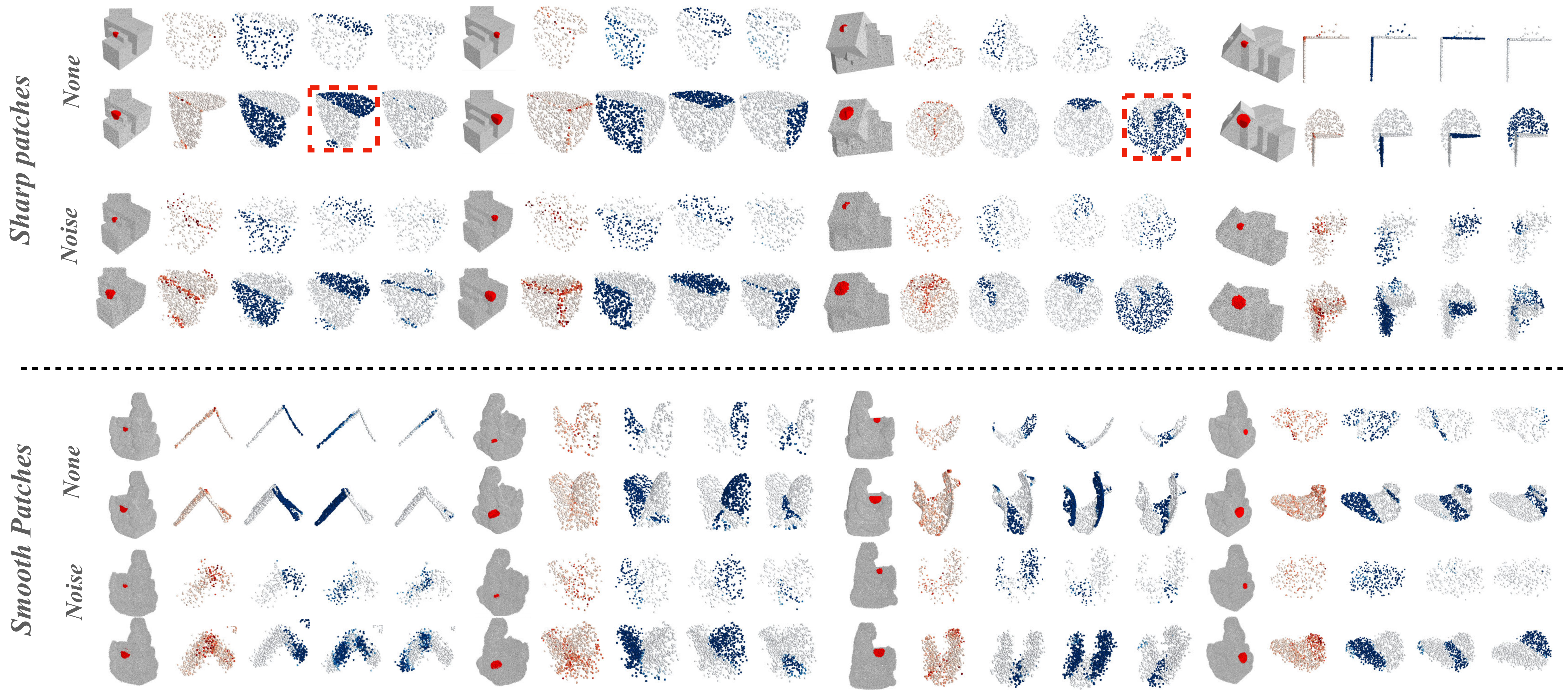}
 \caption{Visualization of angular error and planar experts weights. For each model, from left to right: input model with a specified neighborhood (two sampling scales are given), angular error map, and three planar experts weights. For each kind of model (sharp and smooth), from top to bottom: patches without noise and  0.006 Gaussian noise.}
 \label{fig:3}
\end{figure*}

\subsection{Adaptive local feature aggregation}\label{sec:Features_Aggregation}
For point cloud normal estimation, an accurate pointwise weight prediction can help to softly select the most relevant points and improve normal estimation performance. Inspired by this, we consider calculating pointwise attention weights in a patch, and then we can adaptively extract richer feature at each point by aggregating features of points from its weighted neighborhood. Thus, an adaptive local feature aggregation layer is proposed based on a multi-head point transformer~\cite{vaswani2017attention}. \jun{Fig.~\ref{fig:2} shows some visualization examples in which an average multi-head attention map of the specified point is given. Specifically, given the concatenated feature $F \in R^{K\times C}$ extracted from the basic network, a multi-head scaled dot-product attention is introduced.} In each head, we should first learn a weight matrix (self-attention matrix) $M_{i} \in R^{K\times K}$ between the points:
\begin{equation}
   M_{i} = softmax(\frac{\phi_{i} (F)^{T}\rho_{i}(F)}{\sqrt{K}}),
\end{equation}
where $M_{i}$ is the output attention matrix of the head $i$. $\phi_{i}$ and $\rho_{i}$ are linear projections used to transform pointwise feature.
Then our aggregation features $head_{i}$ can be calculated by the attention matrix as:
\begin{equation}
head_{i} = M_{i}\cdot \alpha_{i}(F),
\end{equation}
where $\alpha_{i}$ is also a linear projection. 
Fig.~\ref{fig:2} qualitatively depicts the average map of the multi-head attentions, which can be calculated by $M = \frac{\sum M_{i}}{n}$ where $n$ is denoted the number of the heads. Finally, the aggregated feature $F_{new}\in R^{K\times C}$ can be obtained:
\begin{equation}
F_{new} =  (head_{1}\bigoplus head_{2} \bigoplus \cdots \bigoplus head_{n})\cdot W,
\end{equation}
where $W$ is a learnable projection matrix and $\bigoplus$ is a concatenation operator. Generally, $n = 8$ in our network.

\begin{table*}[t]
\scriptsize
\caption{Comparison of the RMSE angle error for unoriented normal estimation of our \jun{multi-patch stitching} method (three sampling sizes) to classical geometric methods, and deep learning methods.}
\centering
\begin{tabular}{l ccc ccc ccc c cc ccc}
\toprule
\multirow{2}{*}{} & \multicolumn{3}{c}{\textbf{Ours}}&\multicolumn{3}{c}{\textbf{PCA}} &\multicolumn{3}{c}{\textbf{Jet}}& \multicolumn{1}{c}{\textbf{HoughCNN}}&\multicolumn{2}{c}{\textbf{PCPNET}} & \textbf{Nesti-Net}&\textbf{Lenssen}&\textbf{DeepFit}\\
\cmidrule(r){2-4} \cmidrule(r){5-7}\cmidrule(r){8-10}\cmidrule(r){11-11} \cmidrule(r){12-13} \cmidrule(r){14-14}\cmidrule(r){15-15}\cmidrule(r){16-16}
&1024 &512 &256 &small &med &large
&small &med &large
&ss &ss &ms &ms(MoE) &ss &ss \\
\midrule
None      &7.09   &6.72   &\textbf{6.42}   &8.31  &12.29 &16.77  &7.60  &12.35 &17.35  &10.23  &9.68  &9.62   &6.99   &6.72   &6.51\\
\hline
\textbf{Noise} $\sigma$ \\
0.00125   &9.11   &9.06   &\textbf{9.04}   &12.00 &12.87 &16.87  &12.36 &12.84 &17.42  &11.62  &11.46 &11.37  &10.11  &9.95   &9.21 \\
0.006     &16.24  &\textbf{16.22}  &16.42  &40.36 &18.38 &18.94  &41.39 &18.33 &18.85  &22.66  &18.26 &18.37  &17.63  &17.18  &16.72\\
0.012     &\textbf{21.06}  &21.49  &22.20  &52.63 &27.50 &23.50  &53.21 &27.68 &23.41  &33.39  &22.80 &23.28  &22.28  &21.96  &23.12\\
\hline
\textbf{Density} \\
Gradient  &8.37   &8.09   &7.85   &9.14  &12.81 &17.26  &8.49  &13.13 &17.80  &12.47  &13.42 &11.70  &9.00   &7.73   &\textbf{7.31}\\
Stripes   &7.65   &7.27   &\textbf{6.88}   &9.42  &13.66 &19.87  &8.61  &13.39 &19.29  &11.02  &11.74 &11.16  &8.47   &7.51   &7.92\\
\hline
Average   &11.59  &11.48  &\textbf{11.47}  &21.97 &16.25 &18.87  &21.95 &16.29 &19.02  &16.90  &14.56 &14.34  &12.41  &11.84  &11.80\\
\bottomrule
\end{tabular}
\label{tab:3}
\end{table*}

\begin{table*}[t]
\caption{Comparison of time and space complexity between the proposed approach (different overlapping rate $r$) and other deep learning normals estimation methods.}
\small
\begin{center}
\begin{tabular}{c ccc ccccc}
\toprule
&\textbf{Ours($r$ =12)} & \textbf{Ours($r$ =16)}  &\textbf{Ours ($r$ =24)} & \textbf{DeepFit} &\textbf{IterNet} &\textbf{ Nesti-Net} & \textbf{PCPNet(ms)}\\
\midrule
Num. parameters &19.41M &19.41M  &19.41M & 3.5 M & \textbf{7981} & 170.1 M & 21.3 M\\
Run time & 0.09 &0.13 & 0.20 & 1.02 & \textbf{0.05} & 8.9 & 3.79 \\
Relative time & 1.8$\times$ & 2.6$\times$& 4$\times$ & 20.4$\times$  &\textbf{1 $\times$} & 111$\times$ & 178$\times$ \\
\bottomrule
\end{tabular}
\end{center}
\label{tab:4}
\end{table*}

\subsection{Multi-branch planar experts for normal estimation}\label{sec:Planar_Experts}
After obtaining the pointwise feature $F_{new}$ from the local feature aggregation layer, as shown in Fig.~\ref{fig:1}, a normal estimator module is employed in our patch-level network. An intuitive consideration is that the canonical patch $\hat{P}$ in the 3D space is oriented to three main directions of the axes $x$, $y$ and $z$, the visualization illustrations can be seen in the third to sixth columns of Fig. \ref{fig:3} (see the first model). In this paper, we look for a simple way, namely, \jun{multi-branch planar experts}, to break down these underlying surfaces, thereby reducing the interaction between surface features in the process of normal estimation. Fig.~\ref{fig:3} qualitatively visualizes the performance of multi-planar experts module for different kinds of patches (smooth and sharp ones). 

Specifically, the multi-branch planar experts module has three planar branches and a weight estimation branch. The branch $j$ can give output as $\hat{N}_{j}\in R^{ K\times 3}$ where $j =1,2,3$ and the weight estimation branch learns a probability weight for each branch $w \in R^{K \times 3}$, and $\sum_{j=1}^{3} w_{i, j}=1$ is satisfied for each point $i$ on the patch. The loss function $L$ can be written as:
\begin{equation}
\begin{aligned}
L = \frac{1}{K}\sum_{i=1}^{K} \sum_{j=1}^{3} w(i,j)*\min (&\|\hat{N}_{j}[i]+N[i]\|_{2},\\
&\|\hat{N}_{j}[i]-N[i]\|_{2})
\end{aligned}
\end{equation}
where $N$ and$\hat{N}_{j}$ are the ground truth normal of the patch $P$ and the predicted normal from the branch $j$, respectively. At test time, we compute only one normal, which is associated with the maximal $w$, and the maximum weight of each point will be reused in the patch stitching stage.

\section{Multi-patch normal stitching} \label{sec:Patch_Stitching}
As illustrated in Fig.~\ref{fig:5},  a group of overlapping patches $\{P_{i}\}_{i=1,2,\cdots, M}$ is sampled from model $S$, and a patch stitching operation needs to be executed to realize normal selection from the overlapping patches. Although the number of sampled patches $M \ll N$ (e.g. $10^{3}$ \textit{vs.} $10^{5}$), where $N$ is the point number of model $S$, the total number of points ($K \times M$) is still enormous. The loops that map each point from the overlapping patches to $S$ and complete the multi-normal stitching are time-consuming.

In this paper, the loop process consists of \jun{sparse index matrix} construction and multi-normal selection (see Fig. \ref{fig:1}). First, we perform a sparse matrix of large-scale indicators to realize the unified sorting of the indexes and quickly realize the points \textit{ID}s mapping from the multi-patch to model $S$. Based on the sparse matrix table, we can quickly find the normals and weights from overlapping patches and select the best normal by a weight evaluation process. Specifically, we assume that $\hat{N}_{i} = \{\hat{n}_{i,1},\hat{n}_{i,2},\cdots, \hat{n}_{i,m_{i}}\}$, $\{\hat{p}_{i,1},\hat{p}_{i,2},\cdots, \hat{p}_{i,m_{i}}\}$ and $\{\hat{P}_{i,1}, \hat{P}_{i,2}, \cdots, \hat{P}_{i,m_{i}}\}$ denote the predicted normals from the overlapping parts at point $i$ of model $S$, the correspondence points from the overlapping patches and the correspondence patches respectively, where $m_{i}$ is the number of candidate normals at the point $i$.

The sampled patches have an open boundary, and most of the boundary  points only have neighbors on one side, leading to inaccurate feature extraction. Therefore, we introduce a distance weight to avoid choosing a candidate point far from patch center. The distance weights $ w_{dist}(i,j)$ of each candidate normal $\hat{n}_{i,j}$ is
\begin{equation}
    w_{dist}(i,j) = exp(-\frac{\|\hat{p}_{i,j}-\hat{c}_{i,j}\|_{2}^{2}}{2\sigma^{2}})
\end{equation}
where $j=1, 2, \cdots, m_{i}$ and $\hat{c}_{i,j}$ is the centroid coordinate of patch $\hat{P}_{i,j}$. This means that the best normal should be selected from the point that more near the center of the correspondence patch.   Then, the weights of planar experts  are reused in this stage.  We keep the highest weight from the three experts on the patch and denote it as $w_{pred}(i,j)$. Finally, we can obtain the candidate weights via multiply these two weights as $w_{candidate}(i,j)= w_{dist}(i,j) \cdot w_{pred}(i,j)$. The best normal can be obtained by selecting the maximum candidate weight:
\begin{equation}
\begin{aligned}
    \hat{n}_{selected}(i) &= \hat{N}_{i}[\widetilde{j}],\\
    \widetilde{j} = \mathop{\arg\max}_{j \in \{1,2,\cdots,m_{i}\}} & w_{candidate}(i,j),
\end{aligned}
\end{equation}
where $\hat{n}_{selected}(i)$ denotes the final predicted normal at a point $i$ of model $S$.

\begin{table}[h]
\caption{Architecture of our experts weights estimator.}
\centering
\begin{tabular}{c c c}
\toprule
Layer type & Settings & Outputs\\
\hline
Input &$K\times 192$ & - \\
\hline
Layer 1 &  \tabincell{c}{Conv1d(192,64)\\BatchNorm\\LeakyRelu (0.2)} & $K\times 64$\\
\hline
Layer 2 &  \tabincell{c}{Conv1d(64,32)\\BatchNorm\\LeakyRelu (0.2)} & $K\times 32$\\
\hline
Layer 3 &  \tabincell{c}{Conv1d(32,3)\\Softmax}  & $K\times 3$\\
\bottomrule
\end{tabular}
\label{tab:2-4}
\end{table}

\begin{table}[h]
\caption{Architecture of our planer experts branch.}
\centering
\begin{tabular}{c c c}
\toprule
Layer type & Settings & Outputs\\
\hline
Input &$K\times 192$ & - \\
\hline
Layer 1 &  \tabincell{c}{Conv1d(192,64)\\BatchNorm\\LeakyRelu (0.2)\\Dropout(0.3)} & $K\times 64$\\
\hline
Layer 2 &  \tabincell{c}{Conv1d(64,32)\\BatchNorm\\LeakyRelu (0.2)\\Dropout(0.3)} & $K\times 32$\\
\hline
Layer 3 &  Conv1d(32,3) & $K\times 3$\\
\bottomrule
\end{tabular}
\label{tab:2-3}
\end{table}

\section{Experiments}\label{sec:Experiments}
\subsection{Dataset and training details}
For training and testing, the PCPNet dataset~\cite{guerrero2018pcpnet} is used. The training set consists of eight shapes with different noise levels (no noise, $\sigma$ = 0.00125, $\sigma$  = 0.0065 and $\sigma$  = 0.012), and all shapes are given as triangle meshes and densely sampled with 100k points. The test set consists of 22 shapes, including figurines, CAD objects, and analytic shapes. For evaluation, we use the same 5000 point subset per shape as in Guerrero et al.~\cite{guerrero2018pcpnet}.  In the training stage, we use a batch size of 48, the Adam optimizer, and a base learning rate of $0.1$. Our network is trained on a single Nvidia RTX 2080 Ti GPU.

\jun{In addition, we show our normal estimation branch and experts weights estimator in Tab.~\ref{tab:2-4} and Tab.~\ref{tab:2-3} respectively. We use three normal estimation branches and an experts weights estimator to constitute our multi-branch experts module.}

 \begin{figure}[]
 \centering 
 \includegraphics[width=0.95\columnwidth]{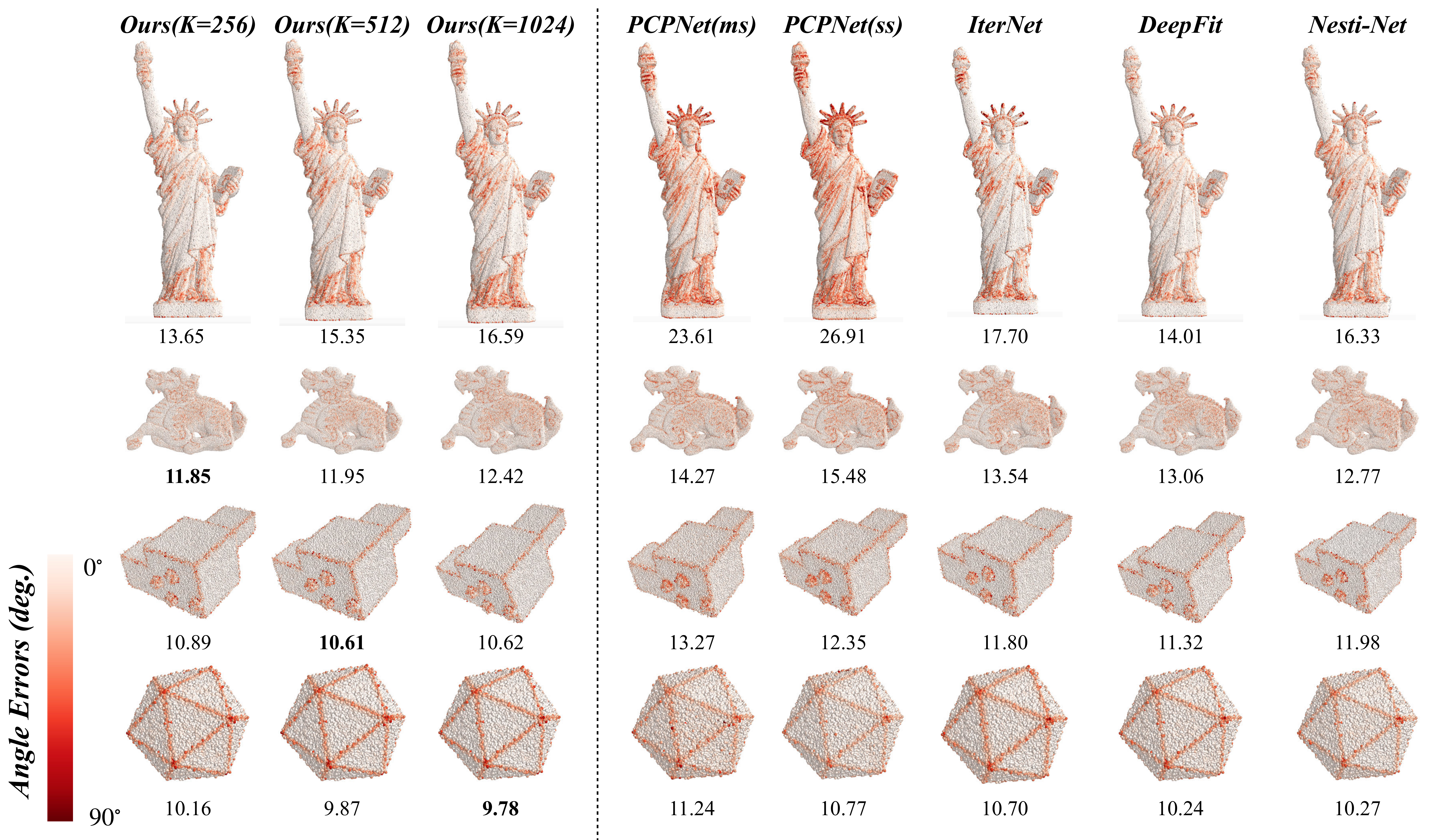}
 \caption{Angular error visualization results of our method (three sampling sizes) compared to others. The colors of the points correspond to angular difference, mapped to a heatmap ranging from 0-90 degrees.}
 \label{fig:4}
\end{figure}

\begin{table*}[h]
\caption{Normal estimation results comparison using the PGP5 metric (higher is better).}
\centering
\scriptsize
\begin{tabular}{l ccc ccc ccc  cc ccc}
\toprule
\multirow{2}{*}{} & \multicolumn{3}{c}{\textbf{Ours}}&\multicolumn{3}{c}{\textbf{PCA}} &\multicolumn{3}{c}{\textbf{Jet}}&\multicolumn{2}{c}{\textbf{PCPNET}} & \textbf{Nesti-Net}&\textbf{Lenssen}&\textbf{DeepFit}\\
\cmidrule(r){2-4} \cmidrule(r){5-7}\cmidrule(r){8-10} \cmidrule(r){11-12} \cmidrule(r){13-13}\cmidrule(r){14-14}\cmidrule(r){15-15}
&1024 &512 &256 &small &med &large &small &med &large &ss &ms &ms(MoE) &ss &ss \\
\midrule
None      & 0.8006  &0.8115   & 0.823  &0.7756  &0.6192 &0.5361  &0.7905  &0.6284 &0.5395  &0.7078  &0.6986  &0.8057   &\textbf{0.8730}   &0.7985 \\
\hline
\textbf{Noise} $\sigma$ \\
0.00125   &\textbf{0.7493}	 &0.7409  &  0.7365 &0.4758 &0.6157 &0.5335  &0.4132 &0.6237 &0.5377  &0.6245  & 0.5932&0.6611& 0.734 & 0.7379 \\
0.006     &	\textbf{0.5818}	&0.5694  & 0.5359   &0.02998 &0.42 &0.4812  &0.027 &0.4152 &0.4837  &0.4486  &0.366 &0.5618  &0.5416 &0.5424\\
0.012     &  \textbf{0.4417}    &0.4172 & 0.3568 &0.0104 &0.154  &0.3719 &0.0099 &0.1462  &0.3715  &0.3156 &0.2482  &0.399  & 0.3634&  0.3132\\
\hline
\textbf{Density} \\
Gradient  &  0.8003 &0.8153   &0.8275 &0.7743  &0.647 &0.4894  &0.7883  &0.6442 &0.4976  &0.6065  &0.6254 &0.7749  &\textbf{0.8778}&0.7912   \\
Stripes   &  0.7641&0.7773   &0.7881&0.7575&0.6174&0.4415 &0.7753  &0.6321  &0.4598 &0.6126  &0.6231  &0.7676 &\textbf{0.8657} & 0.7595\\
\hline
Average   &0.6898  &0.6886  &0.678 &0.4706 &0.5122 &0.4756  &0.4674 &0.5150 &0.4816  &0.5526  &0.5257 &0.6617  &\textbf{0.7092} &0.6572 \\

\bottomrule
\end{tabular}
\label{tab:2-1}
\end{table*}

\begin{table*}[h]
\caption{Normal estimation results comparison using the PGP10 metric (higher is better).}

\centering
\scriptsize
\begin{tabular}{l ccc ccc ccc  cc ccc}
\toprule
\multirow{2}{*}{} & \multicolumn{3}{c}{\textbf{Ours}}&\multicolumn{3}{c}{\textbf{PCA}} &\multicolumn{3}{c}{\textbf{Jet}}&\multicolumn{2}{c}{\textbf{PCPNET}} & \textbf{Nesti-Net}&\textbf{Lenssen}&\textbf{DeepFit}\\
\cmidrule(r){2-4} \cmidrule(r){5-7}\cmidrule(r){8-10} \cmidrule(r){11-12} \cmidrule(r){13-13}\cmidrule(r){14-14}\cmidrule(r){15-15}
&1024 &512 &256 &small &med &large &small &med &large &ss &ms &ms(MoE) &ss &ss \\
\midrule
None      &0.902   &0.9097  & 0.918  &0.8686  &0.7409 &0.6606 &0.8802  &0.7509  &0.6584 &0.8364  &0.8404 &0.9120  &\textbf{0.9288} &0.9074 \\
\hline
\textbf{Noise} $\sigma$ \\
0.00125   &0.8653   &\textbf{0.8667}  & 0.8649 &0.7712 &0.7378&0.6598 &0.7346  &0.7447 &0.6575 &0.8013  &0.8031  &0.8384 &0.8495 &0.8559 \\
0.006     &\textbf{0.7361}  &0.7332  &0.7232  &0.1101 &0.6402 &0.6301  &0.1006 &0.6397 &0.6311  &0.6667  &0.6294 &0.7164  & 0.7154&0.7202\\
0.012     & \textbf{0.6299} &0.6156 & 0.5853 &0.04063 &0.394 &0.5462  &0.0377 &0.3827 &0.547  &0.5546  &0.5124 &0.6123  &0.585 & 0.553\\
\hline
\textbf{Density} \\
Gradient  &0.9093   &0.917  & 0.9239&0.8731  &0.7624 &0.6366  &0.8848  &0.7695 &0.6401  &0.7801  &0.8062 &0.9003  &\textbf{0.9319}  & 0.9114  \\
Stripes   &0.888&0.8956 & 0.9022&0.8609  &0.7379 &0.5879  &0.8743  &0.7504 &0.6001  &0.7967  &0.8076 &0.8929  & \textbf{0.9243} & 0.8888  \\
\hline
Average   & 0.8218 &  \textbf{0.8230}& 0.8212&0.4706 &0.5122 &0.4756  &0.4674 &0.5150 &0.4816  &0.5526  &0.5257 &0.8120  &0.8225  &0.8061 \\

\bottomrule
\end{tabular}
\label{tab:2-2}
\end{table*}

\subsection{Normal estimation performance}
RMSE angle error of our approaches and related methods on the PCPNet test set are presented in Tab.~\ref{tab:3}. 
Compared to HoughCNN~\cite{boulch2016deep} (single-scale) and PCPNet~\cite{guerrero2018pcpnet} (single-scale and multi-scale), Nest-Net~\cite{ben2019nesti} achieves higher accuracy. However, the MoE architecture has 7 sub-networks with a significantly larger number of parameters.  Deepfit~\cite{ben2020deepfit} is based on PCPNet~\cite{guerrero2018pcpnet}. Considering the local latent surface representation, there is a big improvement. However, the time-consuming problem is still not solved. IterNet~\cite{lenssen2020deep} iterates to estimate point normal and have a low RMSE error compared to the above methods, but the network needs to input the whole model simultaneously, which is not a flexible way. In Tab.~\ref{tab:3}, we report results using patch stitching with the following configurations: patch sizes are set to 1024, 512, 256, and patch numbers are set to 2304, 6144, 9216, respectively. Our estimator achieves better performance for the three sampling sizes.  Fig.~\ref{fig:4} also depicts the angular error in each point for the different learning-based methods using a heatmap, and the noise is increasing from top to bottom. It can be seen that large-size sampling can improve the robustness to noise, and our results have consistent advantage for both smooth and sharp models compared to other methods.

\jun{Finally, we also use the proportion of good points metric (PGP $\alpha$), which computes the percentage of points with an error
less than $\alpha$; e.g., PGP10 computes the percentage of points with angular error of less than 10 degrees. Table \ref{tab:2-2} and Table \ref{tab:2-1} report the results of PGP10 and PGP5 respectively for different methods
compared to ours.}

\begin{table}[h]
\scriptsize
\caption{Effects of the basic network, multi-branch planar experts, and adaptive local feature aggregation. The RMSE angle error is used on PCPNet Dataset. The sampling size and patch number are set to 1024 and 1152, respectively.} 
\centering
\begin{tabular}{lcccc}
\toprule
 \textbf{Scale} &\textbf{PCPNet(ss)} &\textbf{ \tabincell{c}{Ours\\(Basic)}} &\textbf{\tabincell{c}{Ours\\(Multi-Branch)} }&\textbf{\tabincell{c}{Ours\\(Multi-Branch\\+Aggregation)}}\\
\midrule
None        &9.68   &8.62   &8.31   &\textbf{7.13}\\
\hline
\textbf{Noise $\sigma$ }                          \\
0.00125     &11.46  &10.62  &10.05  &\textbf{9.18}\\
0.006       &18.26  &17.83  &17.15  &\textbf{16.28}\\
0.012       &22.80  &23.35  &21.85  &\textbf{21.12}\\
\hline
\textbf{Density}\\
Gradient    &13.42  &9.99   &9.38   &\textbf{8.32}\\
Stripes     &11.74  &11.82  &9.57   &\textbf{8.43}\\
\textbf{average} &14.56     &13.70  &12.72  &\textbf{11.74}\\
\bottomrule
\end{tabular}
\label{tab:1}
\end{table}

\subsection{Ablation Study}
\noindent
\textbf{Multi-branch planar experts module.} 
Generally, a sampled patch contains multiple surfaces, see Fig.~\ref{fig:3}. Compare to regressing only the center point normal, patch-level regression is more likely to be affected by the mutual information between the underlying surfaces. The point pairs distributed on different underlying surfaces may interact with each other. Thus, we propose a multi-branch planar experts module to disentangle the interference between the surfaces effectively. Fig.~\ref{fig:3} shows that our planar experts can distinguish the underly surfaces and adaptively adjust the selected surface of each branch according to the numbers of underlying planes (see the red dotted window of Fig.~\ref{fig:3}). Also, see the third column of Tab.~\ref{tab:1},  the \jun{multi-branch planar experts} module gives  a 7.2$\%$ performance boost compared to the basic network (4 layers of residual DGCNN followed only one branch regressor).

\noindent
\textbf{Adaptive multi-scale feature aggregation.} 
Typically, the sampled patches include noise and outliers that heavily reduce accuracy. To overcome this, Deepfit and IterNet learn local weights and fit local surface via weighted least-squares. Differently,  we consider learning patch-level weights in the feature space and adaptively aggregate local features of each point on the patch. Inspired by self-attention network, we introduce an adaptive multi-scale feature aggregation layer. Visualization of the adaptive local weights in different sampling sizes and noise scales are exhibited in Fig.~\ref{fig:2}. It can be seen that the local weights are anisotropic near the edges and corners, assigning high weights on the plane that includes the specified points. Besides, see the red dotted circle of Fig.~\ref{fig:2}, at the convex point, the learned weights drop rapidly with strong self-adaptability. The visualization results also confirm that our aggregation layer is robust to noise (see last row of Fig.~\ref{fig:2}). In Tab.~\ref{tab:3}, compared to the basic network with \jun{multi-branch planar experts module}, the adaptive local feature aggregation layer gives a 7.7 $\%$ improvement dramatically.

\noindent
\textbf{Patch number and sampling size.}
The RMSE error of different sampling sizes and patch numbers are explored (see Tab.~\ref{tab:2}). Firstly, it shows that a large sampling size will improve the high noise cases, and the patch number affects the performance of the network in the density cases. Then, the table also shows that a large patch number will increase the overlapping rate of each point and affects the efficiency of the algorithm. Finally, our accuracy for different parameter choices is relatively stable (from 11.74 to 11.47 on average).

\subsection{Efficiency}
Tab.~\ref{tab:4} lists the number of model parameters and average execution times (ms per point) for estimating normals on a point cloud with 100k points via using an Nvidia GTX 2080 Ti (almost 11G is occupied). IterNet uses a fixed batch of size 100k, and it has the lowest number of parameters and running time. Benefits from patch stitching strategy, the time consumption of our method is not affected by the number of model parameters (Ours has 19.41M parameters but 0.1-0.2 ms/p running time). Our patch-level normal estimation method is the second fast (about 1.8 $\times$ to IterNet) and highly competitive compared with other learning-based methods that estimate normal in a point-by-point manner. 


\begin{table}[htp]
\footnotesize
\caption{The RMSE angle error and average running time with diff. sampling sizes and diff. numbers of patches. The "Overlap" indicates the average number of times that each point participates in the calculation.}
\centering
\begin{tabular}{l cc cc cc}
\toprule
\multirow{2}{*}{} & \multicolumn{2}{c}{\textbf{1024pts}}&\multicolumn{2}{c}{\textbf{512pts}} &\multicolumn{2}{c}{\textbf{256pts}}\\
\cmidrule(r){2-3} \cmidrule(r){4-5}\cmidrule(r){6-7}
\textbf{$\#$ patch} &1152      &2304   &2304   &6144   &4608  &9216 \\
\midrule
None    &7.13     &7.09   &6.84      &6.72   &6.44      &\textbf{6.42}\\
\hline
\textbf{Noise $\sigma$ }  \\
0.00125 &9.18      &9.11   &9.10     &9.06   &9.06    &\textbf{9.04}\\
0.006   &16.28    &16.25  &16.31    &\textbf{16.22} &16.44    &16.42 \\
0.012   & 21.12   &\textbf{21.06}  &21.52    &21.49  &22.27   &22.20 \\
\hline
\textbf{Density}\\
Gradient  &8.32      &7.65   &7.96      &7.27   &7.48      &\textbf{6.88} \\
Stripes   &8.43   &8.37   &8.11     &8.09   &7.92     &\textbf{7.85}\\
\hline
\textbf{average} &11.74       &11.59  &11.64   &11.48  &11.60    &\textbf{11.47}\\
\hline 
\textbf{Times (ms/p)} &0.08    &0.18   &0.09 &0.21   &0.10   &0.19\\
\textbf{Overlap} &12     &24 &12  &32 &12 &24\\
\bottomrule
\end{tabular}
\label{tab:2}
\end{table}

\section{Conclusion}
A fast, accurate, and robust normal estimation framework is proposed, whose efficiency is not affected by network scale. Firstly,  the \jun{adaptive local feature aggregation layer} and \jun{multi-branch planar experts module} are employed to improve the accuracy of the patch-level normal estimator dramatically. Then, in the inference stage, a sparse index matrix is constructed to improve the efficiency of the multi-batch stitching process. Finally, a sufficient weight is used to evaluate the multi-normal of the overlapping parts. The approach demonstrates the competitiveness compared to SOTA approaches.


\bibliographystyle{Bibliography/IEEEtranTIE}
\bibliography{Bibliography/IEEEabrv,Bibliography/BIB_xx-TIE-xxxx}\ 

\end{document}